\documentclass[runningheads]{llncs}

\usepackage{eccv}

\usepackage{eccvabbrv}

\usepackage{graphicx}
\usepackage{booktabs}
\usepackage[table]{xcolor}
\usepackage{amssymb}
\usepackage{subcaption}
\usepackage{wrapfig}
\usepackage{pifont}
\newcommand{\cmark}{\ding{51}}
\newcommand{\xmark}{\ding{55}}
\usepackage[accsupp]{axessibility}  

\usepackage{hyperref}

\usepackage{orcidlink}

\begin{document}

\title{DO-Bench: An Attributable Benchmark for Diagnosing Object Hallucination in Vision-Language Models} 

\titlerunning{DO-Bench}

\author{
JiYang Wang$^*$\inst{1} \and
Jiawei Chen$^*$\inst{1,2} \and
Mengqi Xiao\inst{1} \and
Yu Cheng\inst{1} \and
Yangfu Li\inst{1} \and
Zhaoxia Yin$^\dagger$\inst{1}
}

\institute{
Shanghai Key Laboratory of Multidimensional Information Processing, East China Normal University
\and
Zhongguancun Academy
}

\authorrunning{J. Wang et al.}

\maketitle
\begingroup
\renewcommand\thefootnote{}
\footnotetext{$^*$ Equal contribution.}
\footnotetext{$^\dagger$ Corresponding author.}
\endgroup

\begin{abstract}
Object-level hallucination remains a central reliability challenge for vision–language models (VLMs), particularly in binary object existence verification. Existing benchmarks emphasize aggregate accuracy but rarely disentangle whether errors stem from perceptual limitations or from the influence of contextual textual priors, leaving underlying failure mechanisms ambiguous. We introduce DO-Bench, a controlled diagnostic benchmark that isolates these sources through structured multimodal interventions. Rather than evaluating models in unconstrained settings, DO-Bench probes two complementary dimensions: the Prior-Override dimension progressively strengthens contextual textual priors while holding visual evidence constant to assess resistance to prior pressure, and the Perception-Limited dimension incrementally enhances visual evidence from full-scene context to localized object crops to measure perceptual grounding strength. This paired design enables attribution of errors to prior suppression, perceptual insufficiency, or their interaction. We further define two diagnostic metrics, PriorRobust and PerceptionAbility, to quantify these behaviors consistently. Evaluations across diverse open- and closed-source VLMs reveal systematic differences in prior sensitivity and perceptual reliability, demonstrating that object hallucination reflects heterogeneous, mechanism-dependent failure patterns beyond aggregate accuracy.
\end{abstract}

\section{Introduction}

\begin{table}[t]
\centering
\tiny
\setlength{\tabcolsep}{3.8pt}
\renewcommand{\arraystretch}{1.12}
\caption{Comparison of representative benchmarks for object-level hallucination evaluation. 
We compare whether each benchmark is explicitly designed for object-level hallucination diagnosis, supports binary object existence queries, incorporates controlled test design, enables mechanism-level error attribution, constructs within-image counterfactual instances, and evaluates multiple structured conditions per image.}
\label{tab:benchmark-comparison}
\begin{tabular}{lcccccc}
\toprule
\textbf{Benchmark} &
\textbf{Obj.} &
\textbf{Binary} &
\textbf{Controlled} &
\textbf{Error} &
\textbf{Within-Image} &
\textbf{Multi-Condition} \\
 &
\textbf{Focus} &
\textbf{Questions} &
\textbf{Design} &
\textbf{Attribution} &
\textbf{Counterfactual} &
\textbf{per Image} \\
\midrule
POPE~\cite{li2023evaluating} & \cmark & \cmark & \xmark & \xmark & \xmark & \xmark \\
MME~\cite{fu2023mme} & \xmark & \xmark & \xmark & \xmark & \xmark & \xmark \\
CHAIR~\cite{Rohrbach2018ObjectHI} & \xmark & \xmark & \xmark & \xmark & \xmark & \xmark \\
HallusionBench~\cite{guan2024hallusionbench} & \cmark & \xmark & \xmark & \xmark & \xmark & \xmark \\
HALLUCINOGEN~\cite{seth2025hallucinogen} & \cmark & \xmark & \xmark & \xmark & \xmark & \xmark \\
THRONE~\cite{kaul2024throne} & \cmark & \xmark & \xmark & \xmark & \xmark & \xmark \\
\midrule
\textbf{DO-Bench (ours)} & \cmark & \cmark & \cmark & \cmark & \cmark & \cmark \\
\bottomrule
\end{tabular}
\end{table}


VLMs have achieved remarkable progress in multimodal reasoning, demonstrating strong performance in image captioning, visual question answering, and instruction following\cite{dai2025humanvlm,danish2025comprehensive,ho2025review,lin2025instance}. Despite these advances, object hallucination remains a fundamental reliability bottleneck\cite{zhao2025mitigating,lu2025mitigating,li2023evaluating,hua2025steering}. Models may assert the existence of objects that are absent or deny objects that are clearly present, indicating a breakdown in object-level visual grounding where existence judgments are no longer anchored in visually recognized entities\cite{chen2024multi,zhu2026mmdocbench,lovenia2024negative,li2025mitigating,chen-etal-2025-autobreach,Chen2025ExploringTS}.

Current evaluations of object-level hallucination primarily rely on aggregate accuracy or hallucination rates derived from binary presence judgments. While effective for ranking models, such metrics provide no mechanism-level attribution. False affirmations and false denials are collapsed into a single scalar score, obscuring whether failures stem from perceptual weakness, susceptibility to contextual textual priors, or other confounding factors. As a result, object hallucination is often treated as a uniform phenomenon rather than a composition of distinct decision failures.

\begin{figure}[t]
  \centering
  \includegraphics[width=\linewidth]{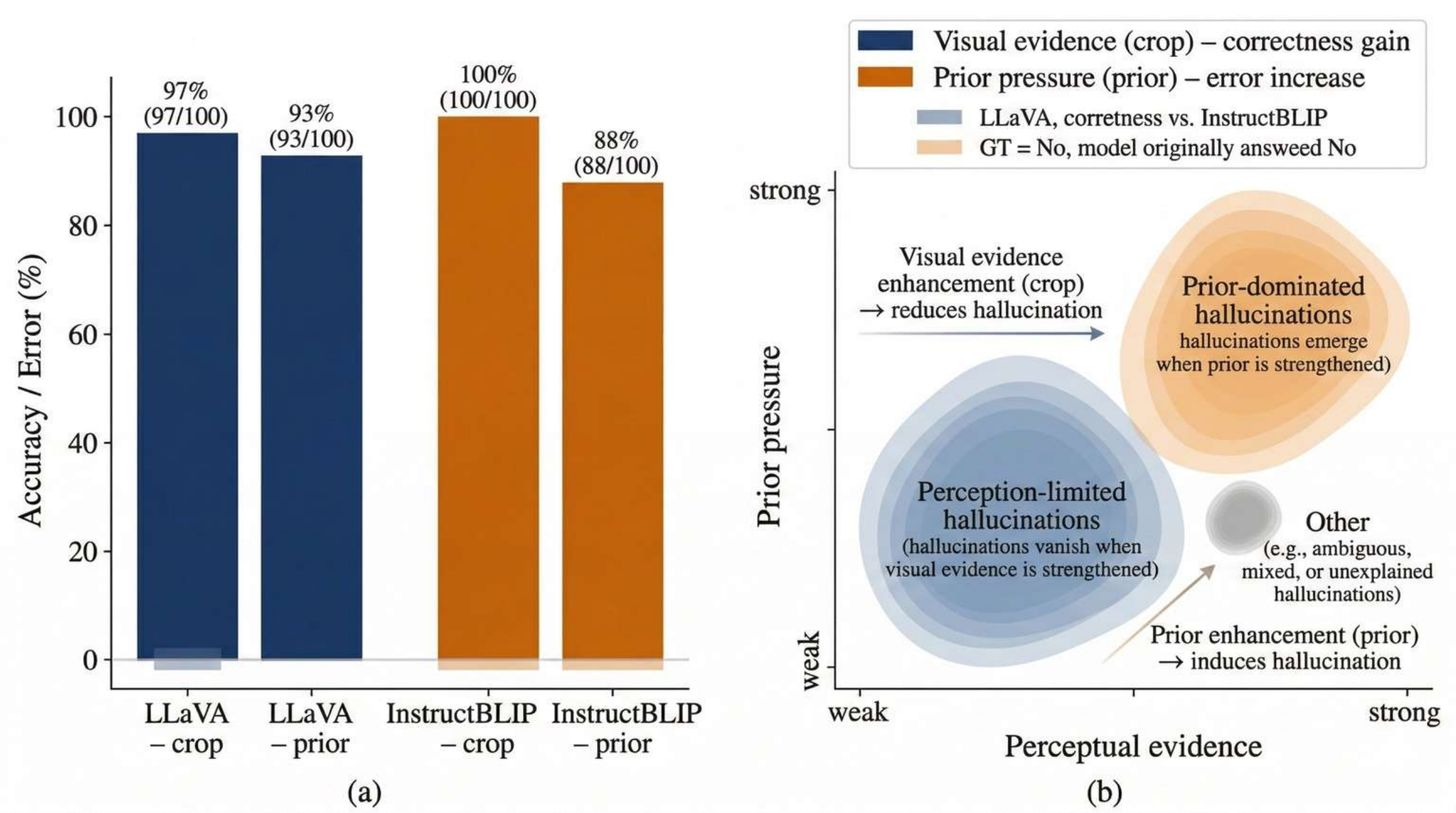}
  \caption{
\textbf{Pilot study.}
Two minimal interventions on object existence verification: target-focused cropping (\textit{crop}) strengthens visual evidence, while contextual prior strengthening (\textit{prior}) increases prior pressure under the same image.
\textbf{(a)} Cropping substantially recovers false denials (correctness gain), whereas stronger priors increase errors by overturning originally correct decisions, for both LLaVA-v1.5-7B and InstructBLIP-7B.
\textbf{(b)} These trends suggest two dominant regimes---\emph{perception-limited} vs.\ \emph{prior-dominated} hallucinations---motivating DO-Bench to disentangle the two factors via controlled within-scene interventions.
}
  \label{fig:pilot}
\end{figure}

A preliminary controlled study further supports this intuition. 
Using LLaVA-v1.5-7B\cite{liu2023visual} and InstructBLIP-7B\cite{dai2023instructblip} on POPE~\cite{li2023evaluating}, we examine object-existence decisions under two minimal interventions: strengthening visual evidence via target-focused cropping, and strengthening contextual prior strength while keeping the image fixed.
As illustrated in Fig.~\ref{fig:pilot}, a large fraction of false denials are recoverable once localized visual evidence becomes more accessible, while a substantial portion of initially correct negative decisions can be overturned when contextual prior strength is sufficiently increased. 
These consistent patterns across models suggest that object-level hallucination is not purely stochastic but systematically modulated by visual evidence accessibility and contextual prior strength.
Further details of the pilot analysis are provided in the Appendix.

Motivated by these observations, we aim to construct a diagnostic benchmark that systematically disentangles object-level hallucination into attributable components. However, such disentanglement is inherently challenging in vision--language settings. Unlike controlled experiments in isolated modalities, visual and linguistic signals are tightly coupled: modifying textual context may alter semantic plausibility, while manipulating visual evidence can introduce scene composition changes or distribution shifts\cite{Wang2023CanLK,gunjal2024detecting,zhou2023analyzing,ye2024multimodal,yang2024language}. As a result, isolating a single causal factor without confounding others requires carefully designed within-image interventions that preserve scene consistency. Achieving reliable attribution therefore demands addressing two intertwined challenges: (i) constructing data that enables controlled variation of prior strength and perceptual accessibility while holding other factors constant, and (ii) defining evaluation metrics that capture structured behavioral trends under intervention rather than collapsing outcomes into aggregate correctness.

To enable controlled data construction, we independently manipulate contextual prior strength and visual evidence strength while keeping the underlying scene unchanged. For prior analysis, we progressively strengthen contextual textual priors while keeping the image fixed, allowing object existence decisions to be observed under increasing contextual prior strength. For perceptual analysis, we construct cases where the target object is present but visually non-salient, and derive hierarchical visual views from the same image, including the full scene, a Cluster view, and a Crop view. These interventions are designed to minimally affect other factors, so observed behavioral changes more directly reflect sensitivity to prior influence or grounding strength.

To enable mechanism-level attribution, evaluation must move beyond static correctness and instead examine how decisions change under controlled variation. If hallucination is driven by contextual textual priors, denial rates should systematically increase as contextual prior strength intensifies. Conversely, if failures arise from weak perceptual grounding, performance should recover when visual evidence becomes more concentrated. Guided by this principle, we introduce mechanism-aware metrics that track structured decision trends under intervention rather than reporting a single aggregate score. By measuring how responses evolve across graded contextual prior strength and strengthened visual evidence, we obtain behavioral signatures that reflect distinct failure mechanisms.

This design allows model differences to be interpreted causally rather than descriptively. Comprehensive evaluation across diverse open- and closed-source VLMs reveals that improvements in perceptual recoverability do not necessarily coincide with increased robustness to contextual-prior-induced override. In particular, scaling enhances recovery under stronger visual evidence but does not uniformly reduce susceptibility to contextual textual priors. These divergent trajectories confirm that object-level hallucination is not a monolithic phenomenon, and that aggregate accuracy alone obscures fundamentally different failure modes.

\paragraph{Contributions.}
\textbf{(1)} We introduce DO-Bench, a controlled diagnostic benchmark for object-level hallucination based on binary object existence verification and structured intervention design.

\textbf{(2)} We develop a data construction protocol that independently manipulates textual prior pressure and visual evidence strength while preserving scene consistency, enabling controlled mechanism isolation.

\textbf{(3)} We propose mechanism-aware evaluation metrics that quantify prior sensitivity and perceptual recoverability through behavioral trend analysis rather than aggregate scoring.

\section{Related Work}

\paragraph{Object hallucination in VLMs.}
Modern vision--language models can produce object existence statements that are weakly supported or unsupported by the image, often due to imperfect grounding and strong language/context priors.
This motivates evaluations that go beyond overall task accuracy and explicitly probe object-level faithfulness\cite{dai2023plausible,zeng2024investigating,zhang2024vision,feng2025vision,lovenia2024negative,jiang2024effectiveness,zhai2023halle,zhao2022vl,parcalabescu2022valse,10.1145/3746027.3755728}.

\paragraph{Benchmarks and evaluation protocols.}
A common paradigm is binary existence probing, where POPE tests whether a model answers Yes/No consistently with image evidence \cite{li2023evaluating}.
For free-form generation, CHAIR quantifies hallucinated object mentions in captions.
Broader suites such as MME and MMMU evaluate multimodal capability at scale, while AMBER expands hallucination assessment across multiple dimensions with automatic scoring \cite{fu2023mme,yue2024mmmu,wang2023amber}.
Diagnostic benchmarks like HallusionBench further stress entangled language hallucination and visual illusion \cite{guan2024hallusionbench,li20251+}.

\paragraph{Mitigation and the attribution gap.}
Training-free mitigation methods include contrastive decoding under visual perturbations (VCD), post-hoc verification and correction (Woodpecker), and decoding-time calibration strategies (OPERA, DoLa) \cite{leng2024mitigating,yin2024woodpecker,huang2024opera,chuang2023dola,li2026deepscan}.
These approaches are typically validated by improvements on overall benchmark scores, making it difficult to determine whether gains come from reducing prior-dominated errors or improving perception under weak evidence.
Our benchmark complements this line by introducing controlled, two-axis interventions that disentangle prior-override from perception-limited failures, enabling attributable diagnosis rather than only ranking.

\section{Do-Bench}
\subsection{Benchmark Design and Task Definition}
\label{sec:ourbench_design}

\begin{figure*}[t]
\centering
\includegraphics[width=\textwidth]{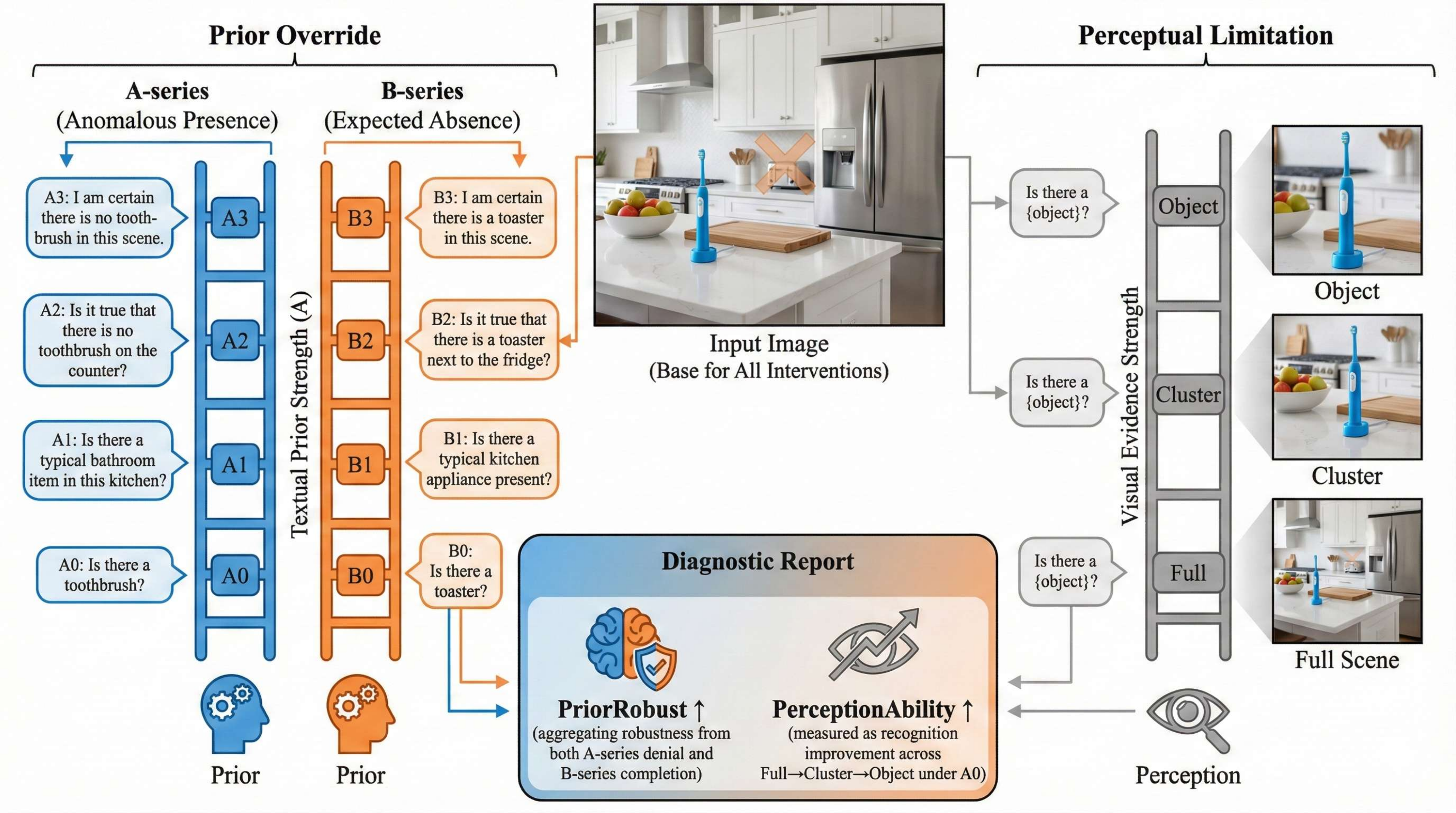}
\caption{
DO-Bench constructs a fixed 10-instance set from each scene via controlled interventions over two dimensions.
Contextual prior strength is varied across four levels for both a present-but-anomalous A-object ($A0$–$A3$) and an absent-but-expected B-object ($B0$–$B3$) under the Full view.
Under neutral contextual priors ($A0$), visual evidence for the A-object is concentrated through Cluster and Crop views.
These instances support two diagnostic readouts: PriorRobust and PerceptionAbility.
}
\label{fig:structure}
\end{figure*}

DO-Bench is organized around \emph{within-scene controls}: multiple evaluation instances are derived from the same scene so that behavioral differences arise from targeted interventions rather than cross-image variation.
This design addresses a key limitation of prior hallucination benchmarks: existence errors may stem from heterogeneous causes (e.g., weak perception or contextual prior influence), yet aggregate scores collapse them into a single scalar without attribution.

\paragraph{Scene-level group structure.}
Each scene forms a fixed \textbf{10-instance group} containing two queried targets under the same visual context.
The \textbf{A-object} is present but contextually unconventional (GT=\emph{Yes}), primarily probing \emph{false denials}.
The \textbf{B-object} is contextually typical but removed (GT=\emph{No}), probing \emph{false affirmations}.
This A/B pairing separates the two major failure directions (denial vs.\ completion) while keeping the surrounding scene unchanged.

\paragraph{Dimension 1: Prior-Override (Full view).}
To isolate contextual prior influence, we vary contextual prior strength across four ordered levels ($0$--$3$) while holding the image fixed (\emph{Full} view).
This yields $A0$--$A3$ for the present A-object and $B0$--$B3$ for the absent B-object.
Since the pixels remain identical, changes in FN/FP across levels can be attributed to strengthened contextual priors.

\paragraph{Dimension 2: Perception-Limited (evidence concentration).}
To probe perceptual limitations, we vary visual evidence accessibility while keeping the prompt fixed at the neutral condition ($A0$).
For the A-object, we derive two additional views from the same scene:
\emph{Cluster}, which retains the object and its immediate context, and
\emph{Crop}, which focuses on the object itself.
Restricting this intervention to $A0$ avoids confounding perceptual recovery with prior influence.

\paragraph{Task and errors.}
Given an image variant and a query about a target object, the model outputs \emph{Yes} or \emph{No}.
Errors decompose into false negatives (denial of present A-objects) and false positives (hallucinated presence for absent B-objects).
By combining graded prior strengthening (Full view) with controlled evidence concentration ($A0$), DO-Bench enables attribution-oriented analysis beyond aggregate leaderboard reporting.

\paragraph{Design rationale.}
The 10-instance within-scene structure provides three advantages:
\textbf{(i) Control:} comparisons are made within the same scene, reducing cross-image variance.
\textbf{(ii) Disentanglement:} prior-driven distortion ($A0$--$A3$, $B0$--$B3$) is separated from perceptual limitation (Full$\rightarrow$Cluster$\rightarrow$Crop).
\textbf{(iii) Aggregation:} structured responses under intervention can be summarized through PriorRobust and PerceptionAbility rather than a single correctness score.

\subsection{Data Construction and Controlled Interventions}
\label{sec:ourbench_construction}

DO-Bench is built through a controlled generation and intervention pipeline that manipulates prior strength and evidence concentration while holding the remaining scene factors fixed.

\textbf{Base scenes (A-series).}
We synthesize base scenes with GPT-image-1.5 using unified prompt templates for ordinary indoor and outdoor environments.
Each scene includes one visually present but contextually unconventional A-object.
To reduce saliency shortcuts, the A-object is placed off-center and embedded in a coherent local cluster, so that successful recognition depends on localized grounding.

\textbf{Absent counterparts (B-series).}
For each scene, we remove a contextually typical B-object via region-specific masking with GPT-image-1.5.
We edit only the target region and avoid global regeneration.
The objective is to preserve geometry and global consistency so that the paired images are intended to differ primarily in object existence.

\textbf{Contextual prior templates and ordinal validation.}
For both A- and B-series, we implement graded contextual prior conflict with a shared four-level template (0–3).
Levels 1–3 differ only by a single certainty token to minimize syntactic and length confounds:
\begin{quote}
A1: I suspect there is no {object} in the image. Is there a/an {object} in the image? Answer Yes or No.\\
A2: I believe there is no {object} in the image. Is there a/an {object} in the image? Answer Yes or No.\\
A3: I am certain there is no {object} in the image. Is there a/an {object} in the image? Answer Yes or No.
\end{quote}

We verify that the four levels form a consistent ordinal ordering across objects and scenes using an NLI-based check with RoBERTa-large-MNLI.
For each level $k$, we compute the contradiction probability between the prior statement and a fixed existence hypothesis:
\begin{equation}
s_k = P(\mathrm{contradiction}\mid \mathrm{contextual\_prior}_k,\ \mathrm{hypothesis}).
\end{equation}
We retain only scene-object pairs that satisfy a predefined monotonicity constraint over $\{s_k\}_{k=0}^{3}$.
These NLI scores are used only for template-level quality control and are not used as supervision for evaluated VLMs.
We additionally validate the intended ordering with human judgments and use two independent LLMs (GPT-5.2 and Qwen3) as a lightweight sanity check to flag phrasing artifacts; both are used only for template cleanliness and ordering validity.

\textbf{Evidence concentration (Cluster/Crop).}
Evidence strengthening is applied only to the A-series under neutral prior ($A0$).
\emph{Cluster} crops a region containing the A-object and its immediate surrounding object cluster, preserving local relations while reducing clutter.
\emph{Crop} isolates the A-object with a consistent margin rule.
Across Full, Cluster, and Crop, only evidence concentration changes; we aim to preserve global scene semantics.

\textbf{Expert verification.}
Five expert annotators review all images, edits, templates, and crops with full-dataset coverage.
They verify target presence/absence, cropping correctness for Cluster/Crop, and ordinal escalation of prior levels.
For B-series pairs, annotators additionally inspect local texture continuity and global coherence to ensure the intervention alters only object existence.
Only samples that pass all checks are retained.
We will release representative before/after examples, prompting specifications, sanity-check procedures, and answer normalization rules for reproducibility and independent inspection.

\textbf{Release.}
DO-Bench, including images, structured annotations, and evaluation code, will be publicly released.

\subsection{Evaluation Protocol and Diagnostic Metrics}
\label{sec:ourbench_eval}

We evaluate all open-source models under deterministic decoding (temperature $=0$) and require binary outputs for all instances.
For API-based systems, we additionally repeat inference for three runs to mitigate potential backend nondeterminism and report averaged results.
We report standard classification metrics (Accuracy, Precision, Recall, F1) for global reference.
To separate prior-driven distortion from perceptual limitation, DO-Bench introduces two diagnostic metrics aligned with the two dimensions.

\textbf{PerceptionAbility} measures visual grounding under the baseline prompt ($A0$) by focusing on the present A-object.
We compute false-negative rates under the two evidence-concentrated views derived from the same base image, Cluster and Crop, and average them to reduce dependence on a single crop rule:
\begin{equation}
\label{eq:perception_ability}
\mathrm{PerceptionAbility} = 100 - \frac{1}{2}\left(FN(A0,\text{Cluster}) + FN(A0,\text{Crop})\right).
\end{equation}
Higher values indicate stronger perceptual grounding under concentrated evidence.

\textbf{PriorRobust} measures robustness to graded contextual prior conflict under both presence and absence conditions.
For the A-series, we compute the area under the false-negative curve over $A0$–$A3$.
For the B-series, we compute the area under the false-positive curve over $B0$–$B3$.
We use trapezoidal integration over levels 0–3, normalize by the maximum possible area, and average the two terms:
\begin{equation}
\mathrm{PriorRobust}=100\cdot\left(1-\frac{1}{2}(\mathrm{A\_AUC}+\mathrm{B\_AUC})\right).
\end{equation}
Higher values indicate greater stability as conflict increases.

Together, PerceptionAbility and PriorRobust provide a compact two dimensional diagnostic summary that complements aggregate accuracy.

\section{Experiment}
We evaluate DO-Bench with two complementary objectives.
First, we assess whether controlled interventions on contextual priors and visual evidence induce structured behavioral changes that cannot be captured by a single aggregate score.
Second, we validate that the proposed attribution metrics provide stable and interpretable summaries under reasonable design variations and sanity checks.
\subsection{Setup}

\begin{figure}[t]
\centering
\begin{minipage}[b]{0.48\textwidth}
\centering
\includegraphics[width=\linewidth]{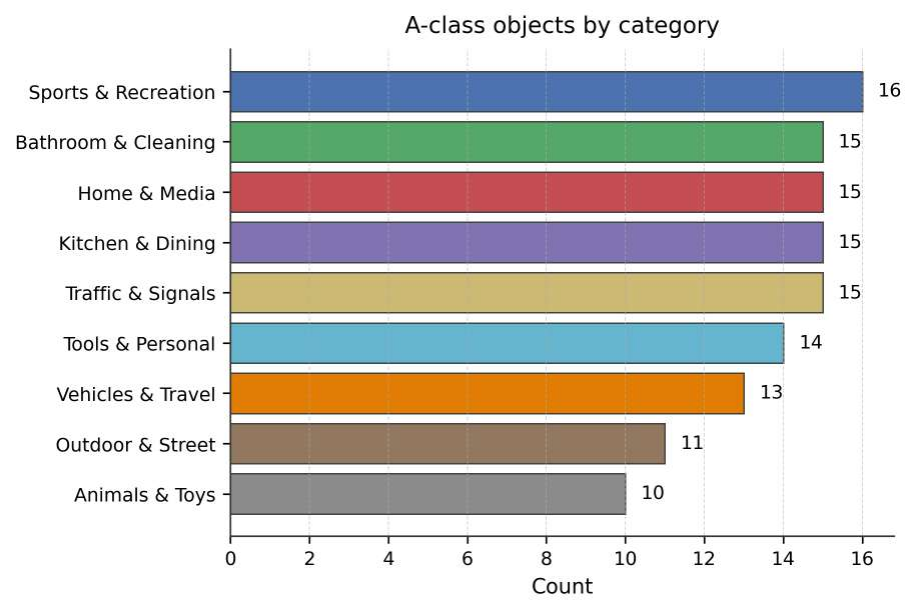}
\subcaption{A-class objects by category}
\label{fig:a_dist}
\end{minipage}
\hfill 
\begin{minipage}[b]{0.48\textwidth}
\centering
\includegraphics[width=\linewidth]{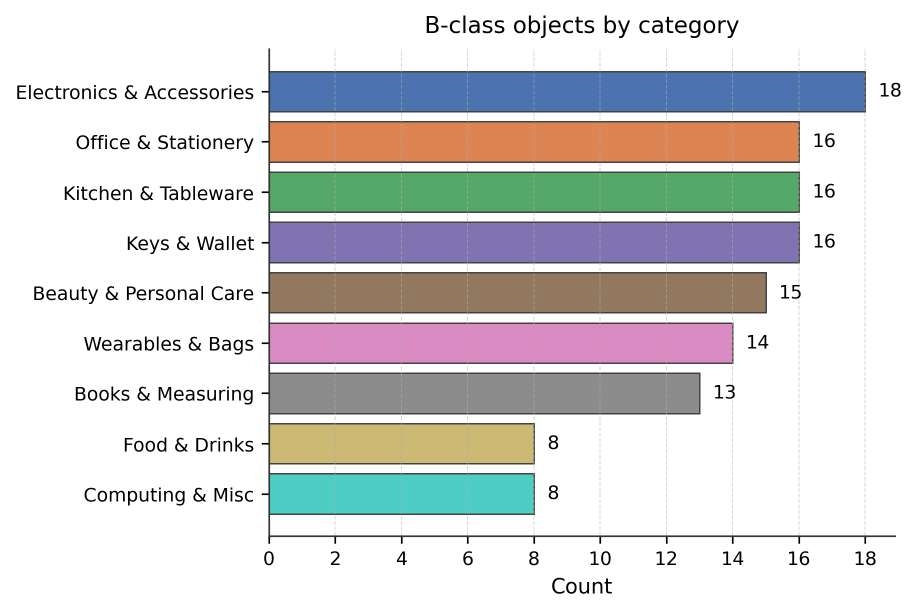}
\subcaption{B-class objects by category}
\label{fig:b_dist}
\end{minipage}
\caption{Object category distribution in DO-Bench. (a) Present A-objects (66 unique categories). (b) Absent B-objects (77 unique categories). Numbers indicate the number of scene groups.}
\label{fig:category_dist}
\vspace{-5mm}
\end{figure}

\paragraph{Dataset.}
DO-Bench consists of 124 structured scene groups constructed under the two-dimensional intervention design described in Section~\ref{sec:ourbench_design}.
Each scene contains one anomalous-but-present A-object (GT=\emph{Yes}) and one expected-but-absent B-object (GT=\emph{No}).
For contextual-prior diagnosis, we instantiate four graded conflict levels under the Full view for each series ($A0$--$A3$ and $B0$--$B3$), yielding 8 instances per scene.
For perceptual diagnosis, we further evaluate the A-series under the baseline prompt ($A0$) with two evidence-concentrated variants (Cluster and Crop).
Each scene therefore yields 10 diagnostic instances, resulting in 1{,}240 total evaluation samples.
Across the 124 scenes, DO-Bench covers \textbf{66 unique A-object categories} and \textbf{77 unique B-object categories}; the corresponding category distributions are shown in Fig.~\ref{fig:category_dist}.
All instances follow a unified yes/no answering protocol with deterministic decoding and answer normalization.
\paragraph{Models.}
We evaluate a diverse set of open- and closed-source vision--language models spanning different architectures and parameter scales.
Open-source models include LLaVA-v1.5 (7B/13B), InstructBLIP-Vicuna (7B/13B), InternVL2.5\cite{chen2024expanding} (2B–78B), and Qwen2.5-VL\cite{Bai2025Qwen25VLTR} (3B–72B).
Closed-source systems include GPT-5.2, Claude-Opus-4.6, and Gemini-3-Flash-Preview-Thinking.
Open-source models are evaluated with deterministic decoding (temperature = 0).
For API-based systems, we perform three independent runs and report averaged results to reduce potential backend stochasticity.
Upon acceptance, we will publicly release the \textbf{API calling code} for closed-source systems together with their corresponding \textbf{official model identifiers} and inference configurations.
Additional implementation details, prompting formats, normalization rules, and prior calibration thresholds are provided in the Appendix.

\begin{table*}[t]
\centering
\tiny
\setlength{\tabcolsep}{2pt}
\renewcommand{\arraystretch}{1.25}
\caption{
Main results on \textbf{DO-Bench}. Performance metrics are reported for reference.
\textbf{PerceptionAbility} measures recognition under enhanced visual evidence under the baseline prompt ($A0$).
\textbf{PriorRobust} aggregates robustness against prior-driven denial (A-series) and prior-driven completion (B-series).
All values are in \%.
}
\label{tab:main_results}
\begin{tabular}{lccccc|cc}
\toprule
& \multicolumn{5}{c}{Performance (\%)}
& \multicolumn{2}{c}{Attribution Metrics (\%)} \\
\cmidrule(lr){2-6} \cmidrule(lr){7-8}
Model
& Acc & Prec & Rec & F1 & Yes Rate
& PerceptionAbility$\uparrow$ & PriorRobust$\uparrow$ \\
\midrule
\multicolumn{8}{l}{\textit{Open-source models}} \\
\midrule
LLaVA-v1.5-7B
& 37.74 & 47.61 & 37.50 & 41.95 & 47.26
& 67.90 & 21.78 \\
LLaVA-v1.5-13B
& 33.47 & 44.34 & 42.61 & 43.45 & 57.66
& 78.29 & 13.98 \\
InstructBLIP-Vicuna-7B
& 28.55 & 40.94 & 43.15 & 42.02 & 63.23
& 89.11 & 8.39 \\
InstructBLIP-Vicuna-13B
& 38.51 & 50.91 & 45.16 & 47.86 & 55.44
& 80.64 & 14.04 \\
InternVL2.5-2B
& 47.58 & 58.25 & 44.62 & 50.53 & 45.97
& 89.12 & 36.70 \\
InternVL2.5-4B
& 56.29 & 82.17 & 34.68 & 48.77 & 25.32
& 66.94 & 49.80 \\
InternVL2.5-8B
& 69.92 & 85.48 & 60.08 & 70.56 & 42.18
& 85.48 & 65.72 \\
InternVL2.5-38B
& 68.15 & 90.30 & 52.55 & 66.44 & 34.92
& 87.90 & 58.87 \\
InternVL2.5-78B
& 72.80 & 95.76 & 59.91 & 73.71 & 39.81
& 87.90 & 63.78 \\
Qwen-VL-7B
& 39.76 & 49.75 & 40.59 & 44.71 & 48.95
& 88.30 & 21.25 \\
Qwen2.5-VL-3B
& 64.11 & 91.64 & 44.22 & 59.66 & 28.95
& 89.12 & 53.36 \\
Qwen2.5-VL-7B
& 65.81 & 96.78 & 44.49 & 60.96 & 27.58
& 81.45 & 56.45 \\
Qwen2.5-VL-32B
& 74.40 & 97.66 & 60.48 & 74.70 & 38.71
& 83.88 & 63.31 \\
Qwen2.5-VL-72B
& 72.31 & 97.39 & 60.08 & 74.31 & 41.13
& 81.05 & 61.62 \\
\midrule
\multicolumn{8}{l}{\textit{Closed-source models}} \\
\midrule
\rowcolor{gray!30}
GPT-5.2
& 85.66 & 97.56 & 80.51 & 88.22 & 55.02
& 87.50 & 84.64 \\
\rowcolor{gray!15}
Claude-Opus-4.6
& 85.22 & 95.16 & 81.99 & 88.09 & 57.44
& 88.71 & 81.45 \\
\rowcolor{gray!45}
Gemini-3-Flash-Preview-Thinking
& 88.53 & 90.53 & 92.47 & 91.49 & 68.10
& 92.74 & 91.45 \\
\bottomrule
\end{tabular}
\end{table*}

\subsection{Main Results}
\label{sec:main_results}

Table~\ref{tab:main_results} summarizes global performance metrics together with PerceptionAbility and PriorRobust.
Three observations are notable.

\paragraph{Aggregate performance does not resolve failure sources.}
Accuracy and F1 provide a coarse ranking, but they conflate qualitatively different errors.
A model may achieve a similar F1 to another model while exhibiting substantially lower PriorRobust, indicating that comparable average correctness can coexist with very different susceptibility to strengthened contextual priors.
Conversely, a model can show strong recoverability under concentrated evidence (high PerceptionAbility) while remaining fragile under prior strengthening (low PriorRobust).
This motivates reporting the two attribution metrics alongside conventional scores.

\paragraph{Perceptual recoverability is often higher than prior robustness.}
Across many model families, PerceptionAbility is relatively high, suggesting that under neutral priors a substantial fraction of present-object denials are recoverable once localized evidence becomes more accessible (Cluster/Crop at $A0$).
In contrast, PriorRobust varies more widely, reflecting heterogeneous sensitivity to strengthened contextual priors across architectures and model families.

\paragraph{Scaling improves overall correctness but does not uniformly improve robustness.}
Within the InternVL2.5 and Qwen2.5-VL families, larger models generally improve Acc/F1, yet improvements in PriorRobust are less uniform.
This indicates that better baseline grounding does not necessarily eliminate prior-induced override effects, and that robustness to prior pressure may be architecture- and training-dependent.

\subsection{Influence of contextual textual priors under controlled strengthening}

\begin{figure*}[t]
\centering
\includegraphics[width=\textwidth]{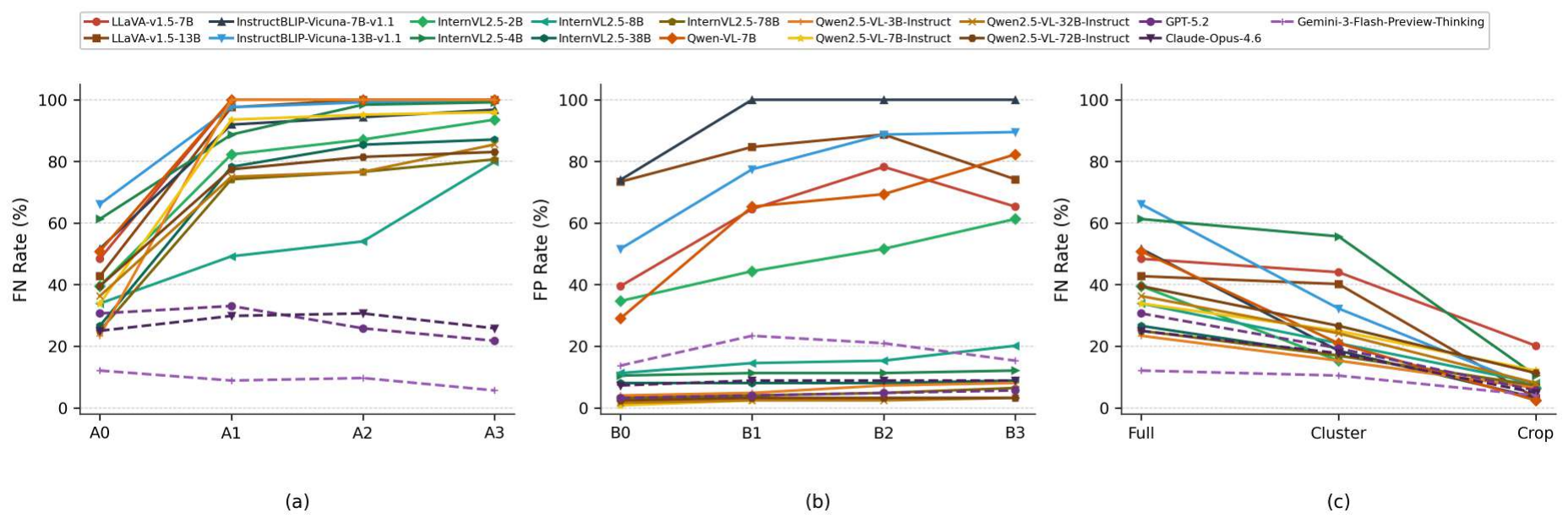}
\caption{
\textbf{Trend-based behavioral analysis on DO-Bench.}
(a) A-series (Prior-Override): FN rate under progressively strengthened contextual prior strength (A0$\rightarrow$A3).
(b) B-series (Prior-Completion): FP rate under increasing contextual prior strength (B0$\rightarrow$B3).
(c) Perception-Limited: FN rate under evidence concentration (Full$\rightarrow$Cluster$\rightarrow$Crop) at the baseline prompt (A0).
Open-source models generally exhibit stronger monotonic amplification under prior strengthening, while closed-source models display lower sensitivity and more stable behavior.
}
\label{fig:trend_analysis}
\vspace{-2mm}
\end{figure*}

Figure~\ref{fig:trend_analysis}(a,b) analyzes how model decisions change when contextual textual priors are progressively strengthened while keeping the image fixed (Full view).

\paragraph{A-series (present objects): prior-induced denials.}
Figure~\ref{fig:trend_analysis}(a) reports FN rates over $A0\rightarrow A3$.
Many open-source models exhibit a clear upward trend: denial errors increase as contextual priors become stronger.
Smaller models often show steeper increases, while larger models tend to start from a lower FN at $A0$ but can still exhibit non-negligible growth under $A2/A3$.
This pattern indicates that improving baseline recognition is not sufficient to prevent prior-driven denial when the textual context becomes strongly conflicting.

\paragraph{B-series (absent objects): prior-driven completion.}
Figure~\ref{fig:trend_analysis}(b) reports FP rates over $B0\rightarrow B3$.
Several open-source models show increased FP under stronger priors, consistent with a prior-driven tendency to complete typical-but-absent objects.
The effect is generally weaker for larger models, but remains noticeable for some architectures.
Closed-source systems maintain lower error rates and typically flatter curves, suggesting higher stability under prior perturbations.

\paragraph{Why we do not require strict monotonicity.}
In practice, model responses may be non-monotonic across $k\in\{0,1,2,3\}$ due to finite sample effects, discrete decoding, and heterogeneous internal heuristics.
DO-Bench therefore treats the curves as empirical response profiles under controlled prior pressure.
PriorRobust summarizes each profile using AUC aggregation, which remains informative even when pointwise monotonicity is imperfect.

Figure~\ref{fig:trend_analysis}(c) evaluates evidence concentration under the neutral prior ($A0$) by comparing Full, Cluster, and Crop views for present A-objects.

\paragraph{Evidence concentration reduces denials across models.}
Across most evaluated models, FN decreases as visual evidence becomes more localized.
This consistent reduction suggests that, under neutral priors, a substantial fraction of denial errors arises from insufficient utilization of localized evidence in cluttered scenes and can be mitigated when the target region becomes more accessible.

\paragraph{PerceptionAbility as recoverability summary.}
PerceptionAbility aggregates FN under Cluster and Crop (Eq.~\ref{eq:perception_ability}), reducing dependence on a single crop rule.
A high PerceptionAbility indicates strong recoverability once evidence is concentrated, while a lower value implies that denials persist even under localized views, pointing to a deeper limitation in perceptual grounding or visual encoding.

\subsection{Orthogonality Between Contextual Prior Strength and Visual Evidence}
\label{sec:prior_evidence_orthogonal}

\begin{figure*}[t]
\centering
\includegraphics[width=\textwidth]{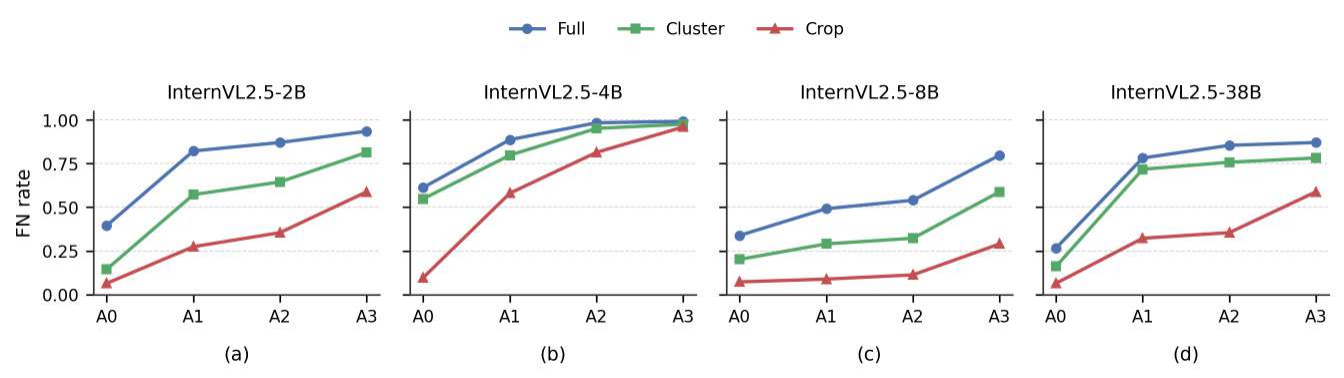}
\caption{
\textbf{Prior--evidence orthogonality in the InternVL2.5 family.}
We jointly vary contextual prior strength ($A0$--$A3$) and visual evidence concentration (Full/Cluster/Crop) for present A-objects, and report the false-negative (FN) rate.
Across model scales, FN increases with stronger contextual priors at fixed evidence, while decreasing as evidence is concentrated at fixed prior, indicating structured and approximately separable effects of the two dimensions.
}
\label{fig:prior_evidence_internvl_family}
\vspace{-5mm}
\end{figure*}

We further investigate whether contextual prior strength and visual evidence strength behave as separable dimensions.
Using the InternVL2.5 family (2B, 4B, 8B, 38B), we construct a $4\times3$ grid for each A-object by jointly varying prior levels ($A0$–$A3$) and evidence conditions (Full, Cluster, Crop).

Along the \emph{prior dimension}, we progressively strengthen contextual textual priors under a fixed image and question structure.
Along the \emph{evidence dimension}, we concentrate visual information under the baseline prompt ($A0$).
Each cell yields a false-negative (FN) rate, producing prior-response curves under three evidence conditions (Fig.~\ref{fig:prior_evidence_internvl_family}).

Across scales, three consistent regularities emerge.
(1) At every fixed prior level, FN decreases monotonically from Full to Cluster to Crop, indicating that denial errors under strengthened contextual priors remain perceptually recoverable.
(2) At every fixed evidence level, FN increases systematically from $A0$ to $A3$, showing that contextual prior strength continues to influence decisions even under concentrated visual evidence.
(3) The three evidence curves are approximately parallel for most models, implying that stronger evidence primarily induces a vertical downward shift of the prior-response curve rather than altering its shape.

InternVL2.5-4B represents a saturation case: under $A3$, FN approaches the upper bound across evidence conditions, reducing recovery margins while preserving the overall trend structure.

Overall, the $4\times3$ analysis demonstrates that visual evidence strength and contextual prior strength operate as approximately separable dimensions, jointly yielding structured and interpretable behavioral shifts.

\subsection{Ablation Studies}
\label{sec:ablation_expanded}

We conduct three ablations to assess metric stability and potential confounds.

\paragraph{Ablation A: Stability of diagnostic metrics.}
We test whether PerceptionAbility and PriorRobust are sensitive to reasonable implementation choices.
For PerceptionAbility, we vary the crop margin (10\%, 20\%) for Cluster/Crop.
For PriorRobust, we compute AUC-based scores using three template families (T1--T3), and report rank consistency.
As shown in Table~\ref{tab:metric_ablation}, PerceptionAbility varies marginally under crop-margin changes (typically within $3\%$ absolute), indicating that recoverability is not driven by a specific crop boundary.
PriorRobust exhibits high rank consistency across template families, with Spearman correlations
$\rho(\text{T1},\text{T2})=0.88$,
$\rho(\text{T1},\text{T3})=0.88$, and
$\rho(\text{T2},\text{T3})=1.00$.
Overall, these results suggest that the attribution metrics are stable under reasonable variations of crop rules and template families.

\paragraph{Ablation B: Editing artifacts vs.\ B-series false positives.}
A potential concern is that localized editing artifacts might spuriously increase B-series false positives.
We estimate an artifact strength score $T$ from before/after differences (difference-based masking, boundary gradient shift, illumination inconsistency), and correlate $T$ with the total number of B-series false positives (B0--B3).
Table~\ref{tab:artifact_ablation} reports Spearman $\rho$ and $p$-values.
All correlations are weak and not statistically significant ($p>0.05$), suggesting that the measured low-level artifact strength alone does not explain the observed B-series FP patterns.
We emphasize that this ablation targets one plausible confound (visible low-level artifacts) and does not fully rule out higher-level semantic inconsistencies; additional analysis is left for future work.

\begin{table*}[t]
\centering

\begin{minipage}[t]{0.485\textwidth}
\centering
\tiny
\setlength{\tabcolsep}{1pt}
\renewcommand{\arraystretch}{1.18}
\caption{Metric stability analysis. PerceptionAbility under different crop margins (10\%, 20\%) and PriorRobust under three AUC formulations (T1--T3). All values are percentages.}
\label{tab:metric_ablation}
\begin{tabular}{lccc|ccc}
\toprule
& \multicolumn{3}{c}{PerceptionAbility}
& \multicolumn{3}{c}{PriorRobust} \\
\cmidrule(lr){2-4} \cmidrule(lr){5-7}
Model
& Base & 10\% & 20\%
& T1 & T2 & T3 \\
\midrule

InternVL2.5-2B
& 89.52 & 87.50 & 90.32
& 36.69 & 41.20 & 45.70 \\
InternVL2.5-4B
& 67.74 & 67.34 & 70.56
& 49.80 & 41.94 & 46.03 \\
InternVL2.5-8B
& 86.29 & 87.90 & 85.08
& 65.73 & 61.69 & 56.38 \\
\midrule

Qwen2.5-VL-3B
& 89.52 & 85.08 & 84.68
& 53.49 & 52.62 & 51.75 \\
Qwen2.5-VL-7B
& 81.85 & 81.85 & 79.03
& 56.45 & 65.19 & 60.89 \\
Qwen2.5-VL-32B
& 83.87 & 82.26 & 77.02
& 63.31 & 71.37 & 62.84 \\
Qwen2.5-VL-72B
& 81.45 & 80.24 & 75.81
& 61.63 & 73.05 & 66.40 \\
\midrule

InstructBLIP-7B
& 89.92 & 87.50 & 86.29
& 8.40 & 9.68 & 10.62 \\
InstructBLIP-13B
& 81.45 & 77.82 & 76.61
& 14.05 & 17.27 & 19.42 \\
\bottomrule
\end{tabular}
\end{minipage}
\hfill
\begin{minipage}[t]{0.485\textwidth}
\centering
\tiny
\setlength{\tabcolsep}{2pt}
\renewcommand{\arraystretch}{1.18}
\caption{Correlation between artifact strength $T$ and B-series false positives. Spearman $\rho$ and corresponding $p$-values are reported. None of the correlations are statistically significant ($p>0.05$).}
\label{tab:artifact_ablation}
\begin{tabular}{lcc}
\toprule
Model & $\rho(T, \mathrm{FP\_count})$ & $p$-value \\
\midrule
Qwen2.5-VL-3B & -0.0699 & 0.4406 \\
Qwen2.5-VL-7B & -0.0289 & 0.7502 \\
Qwen2.5-VL-32B & -0.1289 & 0.1537 \\
Qwen2.5-VL-72B & -0.1718 & 0.0564 \\
\midrule
InternVL2.5-2B & 0.0371 & 0.6823 \\
InternVL2.5-4B & -0.0883 & 0.3292 \\
InternVL2.5-8B & -0.0673 & 0.4579 \\
InternVL2.5-78B & -0.1460 & 0.1058 \\
\midrule
InstructBLIP-7B & -0.1155 & 0.2016 \\
InstructBLIP-13B & -0.1476 & 0.1018 \\
\bottomrule
\vspace{-5mm}
\end{tabular}
\end{minipage}

\end{table*}

\paragraph{Ablation C: Real-image sanity check.}
Because DO-Bench is primarily constructed from generated and locally edited images, we additionally build a small real-image subset from 50 LVIS images.
Following the same 10-instance protocol, we obtain 500 evaluation samples, where A-objects use controlled cropping for evidence concentration and B-objects are removed via localized inpainting.
The resulting trends remain consistent: strengthened priors increase denial/completion errors, while evidence concentration reduces false negatives.
Detailed statistics (invalid rates, per-level curves, and per-model breakdowns) are reported in the Appendix, supporting that the main behavioral signatures are not unique to synthetic scenes.

\section{Conclusion}

We presented DO-Bench, a controlled and attributable benchmark for diagnosing object-level hallucination in vision--language models through binary object existence verification. 
Unlike prior evaluations that mainly report aggregate correctness, DO-Bench derives multiple counterfactual instances from the same scene and applies structured interventions that separately modulate contextual textual prior strength and visual evidence accessibility. 
This design enables mechanism-oriented diagnosis rather than leaderboard-only comparison.

To summarize model behavior under intervention, we introduced two diagnostic metrics: \emph{PerceptionAbility}, capturing recoverability of false denials under evidence concentration, and \emph{PriorRobust}, measuring stability under progressively strengthened contextual priors for both present (A-series) and absent (B-series) targets. 
Experiments across diverse open- and closed-source VLMs reveal a consistent decoupling between perceptual grounding and prior robustness: scaling and stronger vision backbones often improve perceptual recoverability, yet susceptibility to prior-induced override can persist and varies substantially across model families. 
These results suggest that object-level hallucination is heterogeneous and mechanism-dependent, and cannot be reliably characterized by a single aggregate score.

DO-Bench provides actionable signals for model development by localizing failures to prior dominance, perception limitation, or their interaction, and offers a principled testbed for evaluating mitigation strategies under controlled conditions. 
We will publicly release the benchmark, annotations, and evaluation code to facilitate reproducible analysis and future progress toward more reliable and interpretable multimodal systems.

%
%
\bibliographystyle{splncs04}
\bibliography{main}
\end{document}